\pdfoutput=1
\documentclass[11pt]{article}

\usepackage[final]{acl}

\usepackage{times}
\usepackage{latexsym}

\usepackage[T1]{fontenc}
\usepackage[utf8]{inputenc}

\usepackage{microtype}

\usepackage{inconsolata}

\usepackage{graphicx}
\usepackage{enumitem}
\usepackage{pgfplots}
\usepackage{tabularx}
\usepackage{caption}
\usepackage{subcaption}
\usepackage{bbm}
\usepackage{url}
\usepackage{amsmath} 
\usepackage{pifont}
\usepackage{arydshln}
\usepackage{makecell}
\usepackage{booktabs}
\usepackage{multirow}
\usepackage{nicefrac}
\usepackage{algorithm}
\usepackage{algorithmic}
\usepackage{amssymb}

\newcommand{\algorithmicforeach}{\textbf{for each}}
\newcommand{\FOREACH}[2][]{%
\\ \algorithmicforeach\ #2\ \algorithmicdo%
{#1}\begin{ALC@for}}
\makeatother

\usetikzlibrary{shapes.geometric}
\pgfdeclareplotmark{mystar}{
    \node[star,star point ratio=2.25,minimum size=6pt,
          inner sep=0pt,draw=black,solid,fill=red] {};
}

\title{\texttt{LiteSearch}: Efficacious Tree Search for LLM}
\author{Ante Wang$^{1}$, Linfeng Song$^{2}$\thanks{Corresponding Authors}\,, Ye Tian$^{2}$, Baolin Peng$^{2}$, Dian Yu$^{2}$, Haitao Mi$^{2}$,\\
{\bf Jinsong Su}$^{1}$\footnotemark[1]\, and {\bf Dong Yu}$^{2}$ \\
$^{1}$School of Informatics, Xiamen University, China \\
$^{2}$Tencent AI Lab, Bellevue, WA \\
\texttt{\normalsize wangante@stu.xmu.edu.cn,} \texttt{\normalsize lfsong@global.tencent.com,} \texttt{\normalsize jssu@xmu.edu.cn}}

\begin{document}
\maketitle
\begin{abstract}
Recent research suggests that tree search algorithms (e.g. Monte Carlo Tree Search) can dramatically boost LLM performance on complex mathematical reasoning tasks.
However, they often require more than 10 times the computational resources of greedy decoding due to wasteful search strategies, making them difficult to be deployed in practical applications.
This study introduces a novel guided tree search algorithm with dynamic node selection and node-level exploration budget (maximum number of children) calculation to tackle this issue.
By considering the search progress towards the final answer (history) and the guidance from a value network (future) trained without any step-wise annotations,
our algorithm iteratively selects the most promising tree node before expanding it within the boundaries of the allocated computational budget.
Experiments conducted on the GSM8K and TabMWP datasets demonstrate that our approach not only offers competitive performance but also enjoys significantly lower computational costs compared to baseline methods.
\end{abstract}

\section{Introduction}

Mathematical reasoning tasks \cite{amini2019mathqa,cobbe2021training,hendrycks2021measuring,lu2022dynamic} have long been acknowledged as challenging.
These tasks require transforming a question into a sequence of reasoning steps, which are subsequently executed to derive the correct answer.
Recently, large language models (LLMs, \citealt{achiam2023gpt, touvron2023llama, jiang2024mixtral}) have demonstrated remarkable potential in addressing them.
A pivotal approach is the employment of Chain-of-Thought (CoT) prompting \cite{wei2022chain,kojima2022large}, which prompts LLMs to break down a question solution into a sequence of reasoning steps before reaching an answer.
% Building upon CoT, Self-Consistency (SC, \citealt{wang2022self}) assembles answers from multiple reasoning paths using majority voting, resulting in significant boost on final performance.

% \begin{figure}[t]
%     \includegraphics[width=0.45\textwidth]{figures/perf_cost.jpg}
%     \caption{Comparison among ours and typical baselines on GSM8K, where conventional methods are marked as \textcolor{violet}{circle}, and the methods guided by the same trained value model (\S \ref{sec:value}) are marked as \textcolor{teal}{triangle}. We take generated \emph{Tokens (k)} by the LLM as computation costs, thus the lower is better.}
%     \label{fig:perf_cost}
% \end{figure}

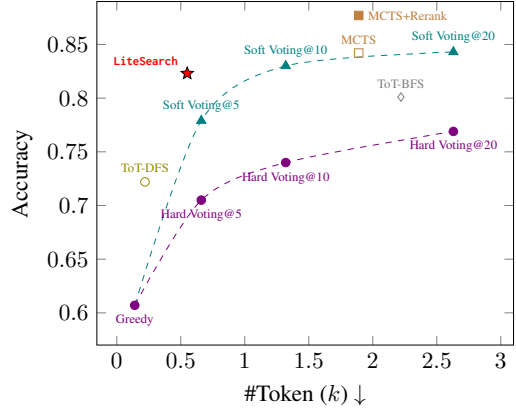
\begin{figure}[t!]
\begin{tikzpicture}[scale=0.8]
\begin{axis}[
    xlabel={\#Token ($k$) $\downarrow$},
    ylabel={Accuracy},
    legend pos=south east,
    legend style={cells={anchor=west}},
    ymin=0.57, ymax=0.89, xmax=3.1,
    ytick={0.6, 0.65, 0.7, 0.75, 0.8, 0.85},
    xtick={0, 0.5, 1, 1.5, 2, 2.5, 3}
]
\addplot [only marks, violet, mark=*] table {
0.14 0.607
2.63 0.769
1.32 0.740
0.66 0.705
};

\addplot [no marks, violet, smooth, thin, dashed] table {
0.14 0.607
0.66 0.705
1.32 0.740
2.63 0.769
};

\addplot [only marks, teal, mark=triangle*, mark size=2.5pt] table {
2.63 0.843
1.32 0.830
0.66 0.779
};

\addplot [no marks, teal, smooth, thin, dashed] table {
0.14 0.607
0.66 0.779
1.32 0.830
2.63 0.843
};

\addplot [only marks, olive, mark=o, mark size=2pt] table {
0.22 0.722
};

\addplot [only marks, gray, mark=diamond, mark size=2pt] table {
2.22 0.801
};

\addplot [only marks, brown, mark=square, mark size=2pt] table {
1.89 0.842
};

\addplot [only marks, brown, mark=square*, mark size=2pt] table {
1.89 0.877
};

\addplot [only marks, red, mark=mystar, mark size=4pt] table {
0.55 0.823
};

\draw[violet,node font=\tiny] node[label={-90:{Greedy}}] at (axis cs:0.14, 0.607) {};
\draw[violet,node font=\tiny] node[label={-90:{Hard Voting@5}}] at (axis cs:0.66, 0.705) {};
\draw[violet,node font=\tiny] node[label={-90:{Hard Voting@10}}] at (axis cs:1.32, 0.740) {};
\draw[violet,node font=\tiny] node[label={-90:{Hard Voting@20}}] at (axis cs:2.63, 0.769) {};

\draw[olive,node font=\tiny] node[label={90:{ToT-DFS}}] at (axis cs:0.22, 0.722) {};
\draw[gray,node font=\tiny] node[label={90:{ToT-BFS}}] at (axis cs:2.22, 0.801) {};
\draw[teal,node font=\tiny] node[label={90:{Soft Voting@5}}] at (axis cs:0.66, 0.779) {};
\draw[teal,node font=\tiny] node[label={90:{Soft Voting@10}}] at (axis cs:1.32, 0.830) {};
\draw[teal,node font=\tiny] node[label={90:{Soft Voting@20}}] at (axis cs:2.63, 0.843) {};
\draw[brown,node font=\tiny] node[label={90:{MCTS}}] at (axis cs:1.89, 0.842) {};

\draw[brown,node font=\tiny] node[label={0:{MCTS+Rerank}}] at (axis cs:1.89, 0.877) {};

\draw[red,node font=\tiny] node[label={145:{\texttt{\textbf{LiteSearch}}}}] at (axis cs:0.55, 0.823) {};
\end{axis}
\end{tikzpicture}
\caption{Comparison among ours and typical systems on GSM8K, where \textit{DFS}, \textit{BFS}, \textit{MCTS} \cite{tian2024toward}\protect\footnotemark and LiteSearch are only guided by the same value network, and \textit{MCTS+Rerank} additionally reranks all searched complete trajectories with value network. \textit{Hard Voting} is self-consistency \cite{wang2022self} and \textit{Soft Voting} additionally use value network to weight each trajectory. We measure the number of generated tokens (\emph{\#Tokens (k)}) by the LLM as computation costs.}
\label{fig:perf_cost}
\end{figure}

Despite their impressive capabilities, LLMs still face challenges when tackling problems with increasing reasoning steps due to the nature of auto-regressive decoding.
This can be analogous to the ``System 1'' mode of thought in psychology \cite{daniel2017thinking}, which is characterized by fast, intuitive, but error-prone thinking.
% \footnotetext{Results from our reimplementation.} % footnote in fig. 1 caption
\footnotetext{The MCTS here does not utilize the guidance from PRM or ORM for fair comparison and efficiency.} % footnote in fig. 1 caption
Much of recent work has focused on enhancing the ``System 1'' capability of LLMs by prompt-engineering, such as hierarchical prompting \cite{suzgun2024meta,zeng2023flowmind} and automatic prompt refine \cite{madaan2024self,yang2024largelanguagemodelsoptimizers,zhou2024self}.
On the other hand, growing research attention is being paid to promote the ``System 2'' mode of thought \cite{daniel2017thinking} for LLMs, which is characterized by deliberative thinking steps with back-and-forth refinements.
These are the key features for solving complex math reasoning tasks.
Particularly, prior efforts have studied enhancing LLMs both at inference time and through self-improvement using tree search algorithms (e.g., DFS and BFS, \citealt{yao2024tree}) and Monte Carlo Tree Search (MCTS, \citealt{feng2023alphazero,tian2024toward,zhang2024accessing,wang2024q}).

However, these approaches often necessitate the creation of expert-designed utility functions \cite{tian2024toward,ma2023let,kang2024mindstar}, making them difficult to be adapted to new scenarios.
Moreover, they are computationally intensive, especially when tackling problems that require numerous logical steps \cite{xie2024self}.
This is because these methods ineffectively manage the expansion budget (the number of nodes to expand) throughout the search process.
As a typical example, BFS adopts a constant budget size throughout the search process, overlooking the fact that some tree nodes do not require much expansion.
Some MCTS approaches \cite{tian2024toward} take adaptive budget based on the importance of each node,
but they still require a large number of simulations or rollouts for accurate statistics to make decisions, and they overlook other important information, such as the depth (progress) of each node.
As the result, there is a pressing need to develop more efficient and adaptable methods for enhancing LLMs' ``System 2'' reasoning capabilities to effectively handle complex reasoning tasks.

In this study,
we introduce a guided tree search algorithm with dynamic node selection and node-level exploration budget calculation, aiming to maintain the performance at a moderate cost.
% Therefore,
Concretely, we employ the value score as guidance to \emph{select} the most promising node for the next action and \emph{expand} it within a dynamically computed budget size, navigating exploration-exploitation balance for guided tree search.
We continue iterating operations of selection and expansion until the resulting trajectory either meets the expected quality score or surpasses the maximum number of iterations.
Notably, the computational budget for each node is inversely correlated to its value score.
This is inspired by the observation that nodes with higher value scores are more likely to yield the correct solution upon expansion, hence we allocate fewer computational resources to them to prevent unnecessary computation and vice versa.
This not only promotes efficient exploitation, facilitating a faster convergence to the final answer, but also guarantees sufficient exploration to cover enough state space for maintaining performance.
% Furthermore, we explore various alternative strategies to enhance these two processes,
% including introducing progress terms when selection to further speed up inference and greedy decoding results during expansion to better estimate budgets.
% They offer more practical alternatives tailored to different requirements.

We conduct experiments on popular GSM8K \cite{cobbe2021training} and TabMWP \cite{lu2022dynamic}. Results show that our methods offer competitive performance but significantly less computation costs (saving around 5$\times$) compared to other baselines.
Detailed analyses confirm the usefulness of each component and provide more practical options for various settings.
Additionally, we also identify the limitations of this research line and suggest possible ways to tackle them.

\section{Related Work}

Thanks to the robust capabilities of LLMs, significant advancements have been made in mathematical reasoning tasks, surpassing traditional approaches that rely on semantic parsing \cite{matsuzaki2017semantic, hopkins2017beyond} or Abstract Syntax Tree (AST) decoding \cite{li2019modeling, qin2021neural, wu2021math}.

Some studies improved the reasoning capabilities of LLMs through further training.
These efforts involved either manually annotating or automatically generating feasible and challenging problems to fine-tune the LLMs \cite{luo2023wizardmath, yu2023metamath, liu2024augmenting, toshniwal2024openmathinstruct}, as well as devising sophisticated techniques, such as reinforcement learning, for efficient training \cite{luo2023wizardmath, wang2023math, lightman2023let, chen2024self}.

Another line of research focuses on inference-time improvement.
Except for the popular self-consistency \cite{wang2022self}, most of these studies treat this task as a tree search problem and investigate various searching algorithms.
\citet{yao2024tree} were the first to introduce Tree-of-Thought (ToT), incorporating Depth-First Search (DFS) and Breath-First Search (BFS) to address reasoning problems.
\citet{khalifa2023grace,zhu2024deductive,xie2024self} applied step-wise Beam Search to math problems, which operates similarly to BFS under certain parameter conditions.
To guide the search process, these studies above either directly prompt the LLMs to evaluate the quality of each step \cite{yao2024tree,xie2024self}, or train a verifier on corresponding datasets to achieve better performance \cite{khalifa2023grace,zhu2024deductive}.

Later research delved into other sophisticated search algorithms, such as Monte Carlo Tree Search (MCTS, \citealt{tian2024toward,zhang2024accessing,wang2024q}), A$^{*}$ \cite{ma2023let}, and Levin Tree Search \cite{kang2024mindstar}.
Nonetheless, these approaches necessitate more robust verifiers to steer the search procedure.
Concretely, \citet{tian2024toward} utilize a blend of the value function, Process-supervised Reward Model (PRM), and Outcome-supervised Reward Model (ORM). \citet{ma2023let} and \citet{kang2024mindstar} train their PRM models on PRM800K \cite{lightman2023let}, which offers manual annotations for 800$k$ reasoning steps of problems from MATH \cite{hendrycks2021measuring}.

This study also follows the same research line, yet it concentrates on developing an efficient algorithm to decrease computation costs while maintaining performance.
Besides, we employ a naive but more practical value network as the verifier, which is trained solely with the final answer labels as distant supervision.

\section{LiteSearch}
\label{sec:method}

\begin{figure*}[t]
    \centering
    \includegraphics[width=0.9\textwidth]{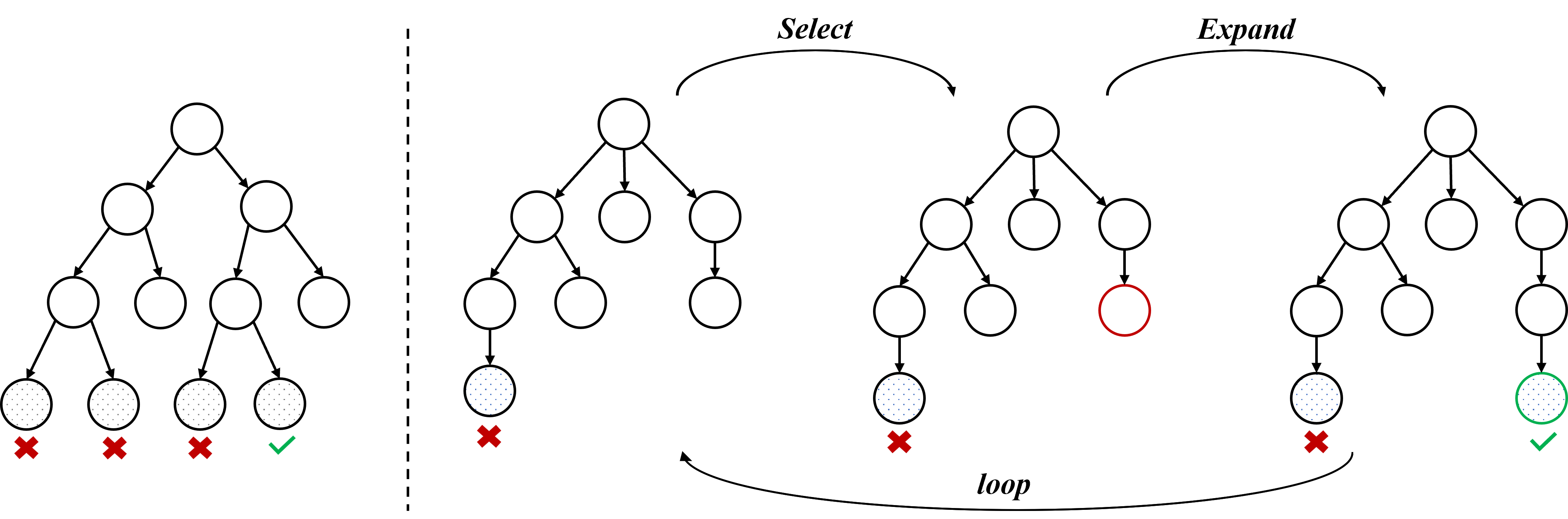}
    \caption{Framework of our guided tree search (Right), which is an iteration of two major steps: \textit{select} and \textit{expand}. Compared with a previous method (BFS, Left), our method dynamically selects the most promising node and expands it within an offered budget, thus reaching the answer without visiting unnecessary nodes.}
    \label{fig:framework}
\end{figure*}

\begin{algorithm}
\caption{LiteSearch}
\begin{algorithmic}[1]
\REQUIRE question $q$, maximum iterations $N$, threshold $\varepsilon$, policy $\pi$, value network $v$
\STATE Initialize tree $\mathcal{T}$ with $q$ as the root
\STATE Set $i \leftarrow 0, \hat{y} \leftarrow \textit{null}$
\WHILE{$i < N$}
    \STATE Select node $s'$ from $\mathcal{T}$ using Eq. \ref{eq:select}
    \STATE Expand $s'$ to obtain its child nodes $\mathbf{C}$ using Eq. \ref{eq:budget}
    \FOR{$c \in \mathbf{C}$}
        \STATE $\mathcal{S} \leftarrow \mathrm{return\_path}(\mathcal{T}, c)$
        \IF{$\mathrm{is\_terminal}(\mathcal{S})$ and $v(\mathcal{S}) > \varepsilon$}
            \STATE Set $\hat{y} \leftarrow \mathcal{S}$
            \STATE \textbf{break}
        \ENDIF
    \ENDFOR
    \STATE $i \leftarrow i + 1$
\ENDWHILE
\ENSURE $\hat{y}$
\end{algorithmic}
\label{alg:search}
\end{algorithm}

\subsection{Guided Tree Search Algorithm}

Taking each math reasoning question $q$ as a tree search problem,
we initialize the root of the search tree with question $q$, while the other tree nodes represent reasoning steps (e.g., $s_i$) generated by an LLM (denoted as policy $\pi$).
Concretely, we treat an (incomplete) trajectory $q, s_1, ..., s_i$ as the state $\mathcal{S}_i$.\footnote{Specially, we define $\mathcal{S}_0 = q$.}
Then, a next step can be sampled from the LLM which consumes $\mathcal{S}_i$:
\begin{equation}
    s_{i+1} \sim \mathtt{LLM}(\mathcal{D}, \mathcal{S}_i),
\end{equation}
where $\mathcal{D}$ is the in-context demonstrations made of question-solution pairs.

As shown in Alg. \ref{alg:search} and Fig. \ref{fig:framework}, our algorithm mainly comprises an iterative process of \textit{Selection} (\S \ref{sec:select}) and \textit{Expansion} (\S \ref{sec:expand}) operations.
For each loop, we first \textit{select} the most promising node, and then \textit{expand} it within the constraints of the computational budget.
Both operations are guided by a value network $v$ (\S \ref{sec:value}).
The algorithm terminates when the generated answers meet the expected value threshold $\varepsilon$ or the number of iterations reaches the limit $N$.

\subsubsection{Selection}
\label{sec:select}

We mainly select the tree node with the highest value for expansion.
Besides, we introduce a \textit{progress term}, denoted as $p(\mathcal{S})$, which quantifies the advancement of a state $\mathcal{S}$ towards the goal within the search trajectory.
By incorporating this term, we prioritize the exploration of nodes that are expected to lead more rapidly to the final answer.
\begin{equation}
    s' = \max_{s_i} (v(\mathcal{S}_i) + \lambda p(\mathcal{S}_i)),
\label{eq:select}
\end{equation}
where $s'$ denotes the selected node, and $\lambda$ is introduced to regulate the impact of the progress term.

It is non-trivial to estimate the progress of a state, thus we introduce an empirical approach based on the trajectory of greedy decoding.
Specifically, we compute the progress by comparing the number of tokens or steps from a given state to those of the corresponding greedy decoding.
For example, when using step number as the metric, a state with $d$ steps has progress of $\nicefrac{d}{\hat{d}}$, where $\hat{d}$ denotes the total number of steps in the greedy decoded trajectory.

\subsubsection{Expansion}
\label{sec:expand}

During the expansion phase, we aim to balance exploitation and exploration by effectively managing the computation budget allocated to the selected node.
Intuitively, an appropriate budget size can promote efficient exploitation, facilitating a faster convergence to the final answer, while also guaranteeing sufficient exploration to cover enough state space for reducing uncertainty.
In line with this spirit, we further explore two strategies preferring either exploitation or exploration:
\textit{Incremental Expansion} and \textit{Batch Expansion}.

\paragraph{Budget Computaton}
We define the allocated budget for a node $\mathcal{S}$ as the maximum number of its children, denoted as $b$, which primarily depends on the value $v(\mathcal{S})$ and depth $d$ of that node.
\begin{equation}
\begin{aligned}
    b &= \min \left(\lceil \frac{\log (1 - \epsilon)}{d \log (1 - v(\mathcal{S}))} \rceil, B\right),
\end{aligned}
\label{eq:budget}
\end{equation}
where $B$ denotes the upper bound of the budget and $\epsilon$ is the expected accuracy, thus a larger $\epsilon$ (e.g., 0.95) encourages more conservative searching. 
Besides, we employ the $\nicefrac{1}{d}$ term, which fosters exploration at the start of searching but encourages exploitation with $d$ increasing to avoid search space explosion.

As the value scores of the preceding search steps usually suffer a larger variance due to inefficient learning of delayed and sparse rewards \cite{sutton2018reinforcement}, confidence estimation of them is relatively not accurate enough.\footnote{This can also be observed in Fig. \ref{fig:calibration}.}
This inevitably influences the computation of suitable budget sizes.
Therefore, we propose to further calibrate value scores using the values of corresponding trajectory from greedy decoding (denoted as $\hat{v}$), especially for the first few steps.
\begin{equation}
\begin{aligned}
    v'({S}) &= \frac{v(\mathcal{S}) + \nicefrac{\hat{v}}{d}}{1 + \nicefrac{1}{d}},
\end{aligned}
\label{eq:greedy}
\end{equation}
where $v'({S})$ represents the enhanced value calibrated by $\hat{v}$ after normalization.
We add $\nicefrac{1}{d}$ term to mainly help the first several steps.

\paragraph{Expansion Strategies}
We propose two expansion strategies that prioritize efficiency and performance, respectively.
\begin{itemize}
    \item \textbf{Incremental Expansion}: This approach incrementally expands one child node after another. If the budget allows, the same node can be reselected until the budget is fully utilized. This method tends to conserve computational resources by carefully managing the budget.
    
    \item \textbf{Batch Expansion}: In contrast, this strategy consumes the entire budget allocated to a node during each iteration, resulting in the generation of multiple child nodes simultaneously. This method broadens the search space for subsequent iterations, potentially leading to the identification of superior nodes and enhancing overall performance.
\end{itemize}

\subsection{Value Network}
\label{sec:value}

The value network \(v(\mathcal{S})\) seeks to approximate the expected cumulative reward starting from state \(\mathcal{S}\) and following a policy \(\pi\) thereafter. This can be represented as \(v(\mathcal{S}) = \mathbb{E}_{\pi} \left[ R_t \mid \mathcal{S}_{t} = \mathcal{S} \right]\), where \(R_t\) is the discounted return starting from state \(\mathcal{S}_{t}\).

Particularly, given a question $q$ and its correct answer $y$ from an expert demonstration dataset.
Each trajectory with reasoning steps (e.g., $s_i$) and final predicted answer $\hat{y}$ is firstly sampled from the LLM (policy \(\pi\)):
\begin{equation}
    s_1,...,s_n,\hat{y} \sim \mathtt{LLM}(\mathcal{D}, q).
\end{equation}
Then, we only take the answer correctness as distant supervision for each reasoning step to train the value network via Mean Squared Error (MSE) loss:
\begin{equation}
    \mathcal{L} = ( v(\mathcal{S}_i) - \mathbbm{I}[y = \hat{y}] )^{2},
\end{equation}
where $\mathbbm{I}$ denotes an indicator function.

In this work, regardless of the policy used, we simply take Llama3-8B\footnote{\url{https://ai.meta.com/blog/meta-llama-3/}} with a regressive head as our value network.
This regressive head is a randomly initialized linear layer, which consumes the hidden state of the last input token and returns a scalar within $[0,1]$:
\begin{equation}
    v(\mathcal{S}_i) = \mathtt{Head}(\mathtt{Llama3}(P, \mathcal{S}_i)[-1]),
\end{equation}
where $P$ is an instruction to help learning. An example is shown in the Fig. \ref{fig:instruction}.

\section{Experiment}
\subsection{Setup}
\paragraph{Dataset}
We conduct experiments on two popular mathematical reasoning datasets:
\begin{itemize}[leftmargin=*]
    \item \textbf{GSM8K} \cite{cobbe2021training}: This dataset comprises 7,473 training and 1,319 testing grade school math word problems that take 2 $\sim$ 8 steps to solve. Solutions primarily involve performing a sequence of elementary calculations using basic arithmetic operations. % ($+ - \times \div$) to arrive at the final answer. 
    \item \textbf{TabMWP} \cite{lu2022dynamic}: This dataset features 38,431 tabular math word problems, presented in either free-text or multiple-choice formats. We focus on the more general free-text category, consisting of 17,315 training questions and 1,000 randomly sampled test questions for evaluation. In contrast to GSM8K, reasoning in TabMWP is based on the provided tables. Additionally, it spans a wider range of domains and supports various answer types.
\end{itemize}

\paragraph{Models and Hyperparameters}
We employ Mixtral-8$\times$7B \cite{jiang2024mixtral} or Llama3-8B as the policy model and train Llama3-8B as the value network.
For the policy models, we adhere to the standard approach of utilizing 8\,/\,4 shots in-context learning for GSM8K\,/\,TabMWP, with a temperature of 0.6.
By default, we set $N, B, \lambda, \varepsilon, \epsilon$ as 100, 10, 0, 0.8, and 0.9, respectively, and investigate other combinations in our analyses.
For the value networks, we sample 8 trajectories per training instance, also with a temperature of 0.6. 
Then, we train the models for 1 epoch across both datasets, employing the AdamW optimizer \cite{loshchilov2017decoupled} with a learning rate of 5e-6 and a linear learning rate scheduler.
Besides, we allocate 5\% of the training instances as a development set to select the optimal checkpoints as value networks.

\paragraph{Evaluation Metrics}
We adopt answer \textit{Accuracy} and the number of generated tokens (\textit{Tokens (k)}) as evaluation metrics for performance and cost, respectively.
It should be noted that we do not take into account the cost of executing value networks.
This is because a value network only performs the regression task, which incurs significantly lower costs compared to the primary generation task.
Besides, it also can be deployed in parallel in practice.

\paragraph{Baselines}
We consider the following baselines:
\begin{itemize}[leftmargin=*]
    \item \textbf{Greedy Decoding}: It intuitively selects the most probable next token at each decoding step.
    \item \textbf{Hard Voting@$K$ (SC,} \citealt{wang2022self}\textbf{)}: Known as self-consistency, which ensembles the answers from multiple sampled solutions as the final answer using majority voting. We sample $K = \{5, 10, 20\}$ times with a temperature of 0.6.
    \item \textbf{ToT-DFS} \cite{yao2024tree}: We implement it by capitalizing on guidance from our trained value network. Specifically, we prune a node if its value score falls below a threshold of 0.5 and limit the maximum number of children to 5 to prevent infinite loops.
    \item \textbf{ToT-BFS\,/\,BeamSearch} \cite{khalifa2023grace,yao2024tree,xie2024self,zhu2024deductive}: These two methods work similarly for this task. Again leveraging our value networks, we ask each node to expand 5 children and only keep 5 nodes with the highest value scores at each depth to avoid search space explosion.
    \item \textbf{Soft Voting@$K$}: It is an enhancement over hard voting by utilizing our value networks. It softly ensembles the answers of different paths by taking their value scores as weights.
\end{itemize}

\begin{table*}[]
\small
\centering
\begin{tabular}{llcccc}
\toprule
& & \multicolumn{2}{c}{GSM8K} & \multicolumn{2}{c}{TabMWP} \\ 
\cmidrule(lr){3-4} \cmidrule(lr){5-6}
& & Accuracy $\uparrow$ & Tokens (\emph{k}) $\downarrow$ & Accuracy $\uparrow$ & Tokens (\emph{k}) $\downarrow$ \\
\midrule
\multirow{11}*{Mixtral-8$\times$7B}
& Greedy Decoding & .607 & 0.14 & .762 & 0.07 \\
& Hard Voting@5 & .705 & 0.66 & .761 & 0.37 \\
& Hard Voting@10 & .740 & 1.32 & .782 & 0.73 \\
& Hard Voting@20 & .769 & 2.63 & .796 & 1.46 \\
\cmidrule(lr){2-6}
& ToT-DFS & .722 & \textbf{0.22} & .822 & \textbf{0.16} \\
& ToT-BFS & .801 & 2.22 & \underline{.861} & 1.45 \\
& Soft Voting@5 & .779 & 0.66 & .811 & 0.37 \\
& Soft Voting@10 & \underline{.830} & 1.32 & .832 & 0.73 \\
& Soft Voting@20 & \textbf{.843} & 2.63 & .847 & 1.46 \\
& Ours (Incremental) & .797 & \underline{0.41} & \textbf{.863} & \underline{0.22} \\
& Ours (Batch) & \underline{.823} & \underline{0.55} & \underline{.854} & \underline{0.29} \\
\midrule
\midrule
\multirow{9}*{Llama3-8B}
& Greedy Decoding & .485 & 0.18 & .659 & 0.08 \\
& Hard Voting@5 & .572 & 0.57 & .680 & 0.42 \\
& Hard Voting@20 & .667 & 2.38 & .698 & 1.68 \\
\cmidrule(lr){2-6}
& ToT-DFS & .676 & \textbf{0.24} & .704 & \textbf{0.19} \\
& ToT-BFS & \underline{.756} & 1.89 & \underline{.787} & 1.35 \\
& Soft Voting@5 & .689 & 0.57 & .747 & 0.42 \\
& Soft Voting@20 & \textbf{.770} & 2.38 & \textbf{.796} & 1.68 \\
& Ours (Incremental) & .731 & \underline{0.46} & \underline{.779} & \underline{0.27} \\
& Ours (Batch) & \underline{.757} & \underline{0.59} & .776 & \underline{0.35} \\
\bottomrule
\end{tabular}
\caption{Main test results. For methods guided by our value networks, we emphasize the best results in \textbf{bold} and the second\,/\,third-best results with \underline{underlining}.}
\label{tab:main}
\end{table*}

\subsection{Development Experiments}

\begin{figure}
\begin{tikzpicture}[scale=0.8]
\begin{axis}[
    xlabel={Step},
    ylabel={Brier},
    legend pos=north east,
    legend style={cells={anchor=west}},
    ymin=0.13, ymax=0.23
]
% \addplot[
%     dashed,
%     color=black,
%     mark=None,
%     ]
%     coordinates {
%         (0, 0.158)
%         (8, 0.158)
%     };
%     \addlegendentry{\cite{wang2024self}}

\addplot[
    color=black,
    mark=o,
    ]
    coordinates {
        (0, 0.212)
        (1, 0.190)
        (2, 0.172)
        (3, 0.158)
        (4, 0.149)
        (5, 0.145)
        (6, 0.143)
        (7, 0.142)
        (8, 0.142)
    };
    % \addlegendentry{Value Network} 

% \addplot[
%     color=purple,
%     mark=square,
%     ]
%     coordinates {
%         (0, 0.212)
%         (1, 0.182)
%         (2, 0.164)
%         (3, 0.151)
%         (4, 0.144)
%         (5, 0.140)
%         (6, 0.139)
%         (7, 0.138)
%         (8, 0.138)
%     };
%     \addlegendentry{w/ cum. score}    
\end{axis}
\end{tikzpicture}
\caption{Reliability of confidence estimation using value scores at different reasoning steps.}
\label{fig:calibration}
\end{figure}
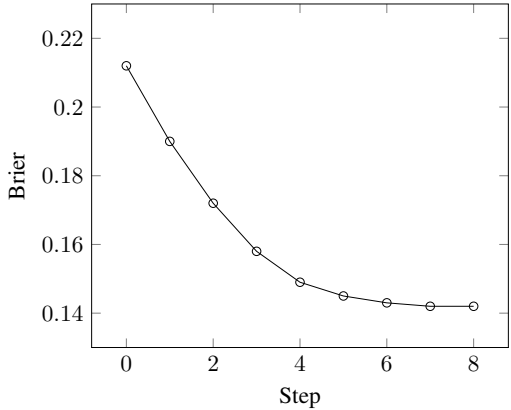

As depicted in Fig. \ref{fig:calibration}, we analyze the alignment between value scores and final correctness.
We employ the widely-used Brier score as our evaluation metric, which is the MSE in effect, indicating that lower scores are preferred.

Generally, as the number of steps increases, we notice a descending trend.
This observation suggests a better correlation between the value scores and the final correctness in subsequent steps.
Remarkably, the Brier scores span from 0.21 to 0.14, marking an improvement of merely 0.07.
This echoes the conclusion of previous research \cite{tian2024toward} and paves the way for our method by taking value scores as estimated confidence of final correctness.

% Compared with \cite{wang2024self}, a sampling-based calibration approach, our value network exhibits competitive performance utilizing just 3 steps.
% Besides, even at the first few steps, our value network can still yield satisfactory calibration results (e.g., 0.18 at the first step).

% We also investigate the cumulative summation of values from the previous steps:
% \begin{equation}
% v_i^{cum} =
% \left
% \{\begin{array}{ll}
% v_i, &i = 0 \\
% \frac{v_i + \gamma v_{i-1}^{cum}}{1 + \gamma}, &i > 0
% \end{array}
% \right.
% \end{equation}
% where $\gamma$ is a hyperparameter to balance the current and previous values. Fig. \ref{fig:calibration} demonstrates that this can further boost the calibration results.

% In light of the aforementioned findings, we inherently treat the value score of a state as the expected confidence of correctness thereafter.
% This paves the way for our guided tree search method, as detailed in Section \ref{sec:method}, which dynamically manages the computational budget to achieve a balance between performance and cost.

\subsection{Main Results}

Table \ref{tab:main} shows the main test results on GSM8K and TabMWP. We observe the following conclusions:

\paragraph{\textit{Value Guidance Boosts Model Performance}}
In line with prior research \cite{wang2022self}, Hard Voting significantly improves Accuracy. However, its costs also proportionately increase with the growing of sampling size $K$.
With the guidance of our value networks, both Soft Voting and tree search algorithms can further enhance Accuracy without incurring additional costs.
Besides, \textit{Soft Voting@5} consistently surpasses \textit{Hard Voting@20}, substantiating the effectiveness of verification as previously discussed in \cite{cobbe2021training}.

\paragraph{\textit{Current Tree Search Algorithms Neglect the Performance-Cost Tradeoff}}
Previous methods, \textit{ToT-DFS} and \textit{ToT-BFS}, prefer different evaluation metrics.
Among the value-guided approaches, \textit{ToT-DFS} consistently has the lowest cost but achieves suboptimal performance.
This is because \textit{ToT-DFS} focuses mainly on pruning bad nodes and lacks the flexibility to select better nodes for further improvement.
In contrast, \textit{ToT-BFS} tackles this shortcoming of \textit{ToT-DFS} by maintaining a branch of nodes with the highest values, thereby resulting in better performance. However, it also unnecessarily visits lots of nodes during the search, leading to significantly higher costs.

\paragraph{\textit{Dynamic Selection and Expansion Maintain Performance and Decrease Cost}}
By fully utilizing the guidance from value networks, our methods achieve the best tradeoff between performance and cost.
Our approach falls within the cost range of \textit{ToT-DFS} and \textit{Soft Voting@5}, yet yields significantly better performance.
For the two expansion strategies, \textit{Incremental} saves nearly 20\% of costs of \textit{Batch} and performs even better on TabMWP.
However, \textit{Incremental} performs noticeably worse than \textit{Batch} on Accuracy on GSM8K, with a 2.6-point lower score.
This is due to \textit{Batch} providing a better comparison among nodes for selection by expanding more nodes each time.
It is worth noting that both of our methods often cannot outperform \textit{Soft Voting@20} on Accuracy. We will provide detailed analyses in \S \ref{sec:error}.

\subsection{Ablation Study and Analyses}

\begin{table}[]
\small
    \centering
    \begin{tabular}{lcc}
    \toprule
    & Accuracy $\uparrow$ & Tokens (\emph{k}) $\downarrow$ \\
    \midrule
    Incremental & \textbf{.797} & 0.41 \\
    ~~~~$\Rightarrow$ static budget & .779 & 0.67 \\
    % ~~~~$\Rightarrow$ cumsum. value & .781 & 0.45 \\
    ~~~~w/o depth penalty & .780 & 0.43 \\
    ~~~~w/o greedy value & .783 & \textbf{0.40} \\
    \hdashline
    Batch & \textbf{.823} & \textbf{0.55} \\
    ~~~~$\Rightarrow$ static budget & .802 & 1.79 \\
    % ~~~~$\Rightarrow$ cumsum. value & .812 & 0.59 \\
    ~~~~w/o depth penalty & .815 & 0.79 \\
    ~~~~w/o greedy value & .806 & 0.62 \\
    \bottomrule
    \end{tabular}
    \caption{Ablation study on dynamic budgeting.}
    \label{tab:ablation}
\end{table}

\paragraph{\textit{Dynamic Budgeting Helps Both Performance and Cost-Efficiency}}
We first study the effectiveness of the dynamic budget size $b$, which is decided by Eq. \ref{eq:budget}.
The following variants are considered:
(1) \textit{$\Rightarrow$ static budget}: We directly set $b$ as $B$, resulting in each node being expanded with a fixed budget size;
(2) \textit{w/o depth penalty}: We remove the $\nicefrac{1}{d}$ term from Eq. \ref{eq:budget}, which previously penalized $b$ as the depth $d$ increased;
(3) \textit{w/o greedy value}: We do not consider Eq. \ref{eq:greedy} to calibrate value scores with greedy results.

As shown in Table \ref{tab:ablation},
we observe that dynamic budgeting helps in both \textit{Incremental} and \textit{Batch} by allowing them to maintain higher accuracy with fewer tokens compared to all other variants.
Specifically, \textit{$\Rightarrow$ static budget} severely hurts both performance and cost, particularly leading to 3 times computation costs when using \textit{Batch Expansion}.
\textit{w/o depth penalty} and \textit{w/o greedy value} perform competitively for \textit{Incremental}, but still have considerable negative influence on \textit{Batch}.
These results highlight the importance of dynamic budgeting especially in scenarios where \textit{Batch Expansion} is employed.

\begin{figure}
\begin{tikzpicture}[scale=0.8]
\begin{axis}[
    xlabel={\#Token ($k$)},
    ylabel={Accuracy},
    legend pos=south east,
    legend style={cells={anchor=west}},
    ymin=0.55, ymax=0.85
]
\addplot[
    color=blue,
    mark=o,
    ]
    coordinates {
        (0.145, 0.607)
        (0.192, 0.690)
        (0.327, 0.764)
        (0.396, 0.786)
        (0.412, 0.797)
    };
    \addlegendentry{w/ Incremental} 

\draw[blue,node font=\small] node[label={145:{$2$}}] at (axis cs:0.192, 0.690) {};
\draw[blue,node font=\small] node[label={145:{$3$}}] at (axis cs:0.327, 0.764) {};
\draw[blue,node font=\small] node[label={145:{$5$}}] at (axis cs:0.396, 0.786) {};
\draw[blue,node font=\small] node[label={90:{$10$}}] at (axis cs:0.412, 0.797) {};

\addplot[
    color=violet,
    mark=square,
    ]
    coordinates {
        (0.145, 0.607)
        (0.215, 0.697)
        (0.347, 0.763)
        (0.424, 0.802)
        (0.554, 0.823)
    };
    \addlegendentry{w/ Batch}   

\draw[violet,node font=\small] node[label={0:{$2$}}] at (axis cs:0.215, 0.697) {};
\draw[violet,node font=\small] node[label={-90:{$3$}}] at (axis cs:0.347, 0.763) {};
\draw[violet,node font=\small] node[label={-45:{$5$}}] at (axis cs:0.424, 0.802) {};
\draw[violet,node font=\small] node[label={-90:{$10$}}] at (axis cs:0.554, 0.823) {};

\draw[black,node font=\small] node[label={-90:{$1$}}] at (axis cs:0.145, 0.607) {};
\end{axis}
\end{tikzpicture}
\caption{Performance of \textit{Incremental} and \textit{Batch} on GSM8K when using different budget limitations $B$, where $B = \{1, 2, 3, 5, 10\}$.}
\label{fig:budget}
\end{figure}
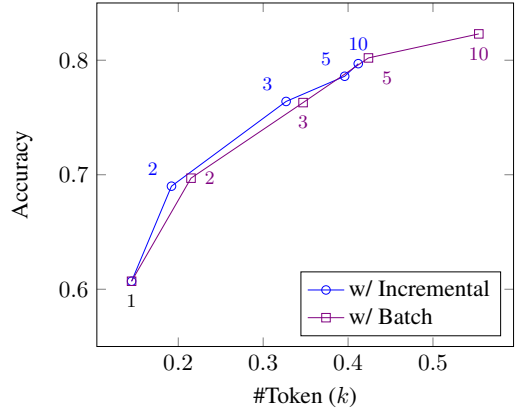

\paragraph{\textit{Influences of Budget Limitation}}
Budget limitation $B$ decides the upperbound of budget size $b$.
As illustrated in Fig. \ref{fig:budget}, we observe a clear tradeoff between performance and cost.
With the growth of $B$, the computation cost also increases correspondingly because larger budget sizes are allocated to challenging states with lower value scores.
Consequently, more problems are correctly solved due to more comprehensive searching.
Regarding the two expansion strategies, \textit{Incremental} perform slightly better than \textit{Batch} with competitive accuracy but fewer costs when $B \leq 3$.
This is because it may not use up all budgets when good nodes have been generated during incremental expansion.
However, \textit{Batch} yields better accuracy by taking more costs when $B=\{5, 10\}$ because it fully utilizes allocated budgets, thus providing larger search space for better selection.

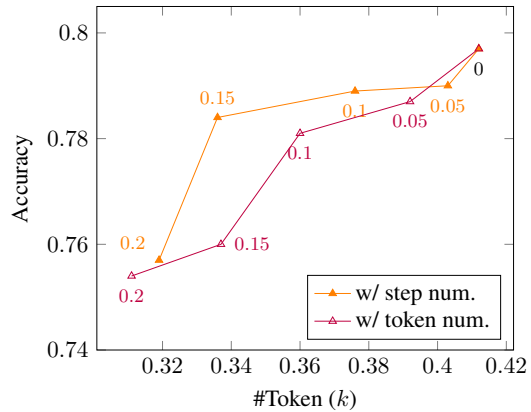
\begin{figure}[t!]
\begin{tikzpicture}[scale=0.8]
\begin{axis}[
    xlabel={\#Token ($k$)},
    ylabel={Accuracy},
    legend pos=south east,
    legend style={cells={anchor=west}},
    ymin=0.74, ymax=0.805
]
\addplot[
    color=orange,
    mark=triangle*,
    ]
    coordinates {
        (0.319, 0.757)
        (0.336, 0.784)
        (0.376, 0.789)
        (0.403, 0.790)
        (0.412, 0.797)
    };
    \addlegendentry{w/ step num.} 

\draw[orange,node font=\small] node[label={145:{$0.2$}}] at (axis cs:0.319, 0.757) {};
\draw[orange,node font=\small] node[label={90:{$0.15$}}] at (axis cs:0.336, 0.784) {};
\draw[orange,node font=\small] node[label={-90:{$0.1$}}] at (axis cs:0.376, 0.789) {};
\draw[orange,node font=\small] node[label={-90:{$0.05$}}] at (axis cs:0.403, 0.790) {};

\draw[black,node font=\small] node[label={-90:{$0$}}] at (axis cs:0.412, 0.797) {};

\addplot[
    color=purple,
    mark=triangle,
    ]
    coordinates {
        (0.311, 0.754)
        (0.337, 0.760)
        (0.360, 0.781)
        (0.392, 0.787)
        (0.412, 0.797)
    };
    \addlegendentry{w/ token num.}    

\draw[purple,node font=\small] node[label={-90:{$0.2$}}] at (axis cs:0.311, 0.754) {};
\draw[purple,node font=\small] node[label={0:{$0.15$}}] at (axis cs:0.337, 0.760) {};
\draw[purple,node font=\small] node[label={-90:{$0.1$}}] at (axis cs:0.360, 0.781) {};
\draw[purple,node font=\small] node[label={-90:{$0.05$}}] at (axis cs:0.392, 0.787) {};

\end{axis}
\end{tikzpicture}
\caption{Performance of \textit{Incremental Expansion} on GSM8K when using \textit{step number} and \textit{token number} to estimate progress term $p(\mathcal{S})$, where $\lambda = \{0, 0.05, 0.1, 0.15, 0.2\}$.}
\label{fig:progress}
\end{figure}

\paragraph{\textit{Influence of Progress Estimation}}
We then investigate the choice of $p(\mathcal{S})$ and $\lambda$ in Eq. \ref{eq:select}.
We consider \textit{step number} and \textit{token number} against corresponding results of greedy decoding to estimate $p(\mathcal{S})$.
As depicted in Fig. \ref{fig:progress}, increasing $\lambda$ improves cost-efficiency by prioritizing nodes with faster progress at the risk of inaccuracy.
Comparing \textit{step number} and \textit{token number}, the former is relatively better with a modest downward trend.
By sacrificing 1.3 points in accuracy, utilizing \textit{step number} and $\lambda=0.15$ saves nearly 20\% computational costs.
In contrast, the efficacy of \textit{token number} is unsatisfactory. 
This can be attributed to its higher degree of variability, thus yielding less precise estimates of progress.

\begin{figure}
\begin{tikzpicture}[scale=0.8]
\begin{axis}[
    xlabel={Difficulty},
    ylabel={Tokens ($k$)},
    legend pos=north west,
    legend style={cells={anchor=west}},
]
\addplot[
    color=blue,
    mark=o,
    ]
    coordinates {
        (0.0, 0.117)
        (0.1, 0.145)
        (0.2, 0.189)
        (0.3, 0.211)
        (0.4, 0.21)
        (0.5, 0.389)
        (0.6, 0.393)
        (0.7, 0.513)
        (0.8, 0.713)
        (0.9, 1.084)
        (1.0, 1.096)
    };
    \addlegendentry{w/ Incremental} 
\addplot[
    color=violet,
    mark=square,
    ]
    coordinates {
        (0.0, 0.146)
        (0.1, 0.215)
        (0.2, 0.245)
        (0.3, 0.32)
        (0.4, 0.324)
        (0.5, 0.487)
        (0.6, 0.575)
        (0.7, 0.753)
        (0.8, 0.926)
        (0.9, 1.242)
        (1.0, 1.686)
    };
    \addlegendentry{w/ Batch} 
\addplot[
    dashed,
    color=black,
    mark=none,
    ]
    coordinates {
        (0.0, 0.091)
        (0.1, 0.099)
        (0.2, 0.117)
        (0.3, 0.123)
        (0.4, 0.125)
        (0.5, 0.146)
        (0.6, 0.131)
        (0.7, 0.141)
        (0.8, 0.185)
        (0.9, 0.17)
        (1.0, 0.161)
    };
    \addlegendentry{Greedy}   
\end{axis}
\end{tikzpicture}
\caption{Cost of \textit{Incremental} and \textit{Batch} with the growth of difficulty on GSM8K. \textit{Difficulty} is defined as ``$1-x$'', where $x$ is the frequency of gold answer in 20 sampled paths.}
\label{fig:diff_cost}
\end{figure}
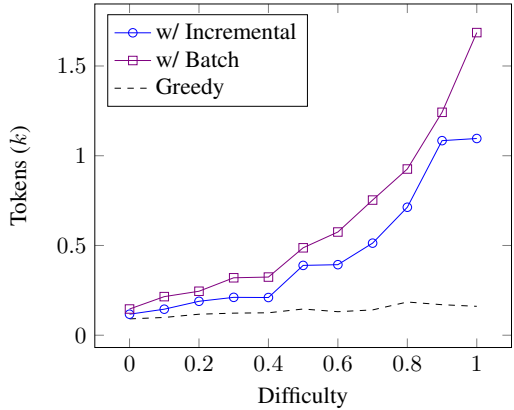

\paragraph{\textit{Harder Problems are Allocated Larger Budgets}}
Fig. \ref{fig:diff_cost} illustrates the correlation between cost and question difficulty.
Inspired by \cite{wang2024self}, we estimate the difficulty of a question by computing the frequency of the gold answer in multiple sampled paths after inversion (``$1-x$'').
We observe that for easier questions, our methods cost competitively to \textit{Greedy Decoding}.
However, as the difficulty escalates, the cost of our method also rises proportionately.
Regarding our expansion strategies, \textit{Batch} consistently takes higher costs and the gap also widens with the difficulty increases.

\begin{figure}
\begin{tikzpicture}[scale=0.8]
\begin{axis}[
    xlabel={Ratio of Data},
    ylabel={Accuracy},
    legend pos=south east,
    legend style={cells={anchor=west}},
]
\addplot[
    color=teal,
    mark=square*,
    ]
    coordinates {
    (0.01, 0.75)
    (0.05, 0.792)
    (0.1, 0.808)
    (0.2, 0.821)
    (0.3, 0.834)
    (0.5, 0.851)
    (1.0, 0.863)
    };
    \addlegendentry{Single}  
\addplot[
    color=cyan,
    mark=halfsquare right*,
    ]
    coordinates {
    (0.01, 0.779)
    (0.05, 0.815)
    (0.1, 0.815)
    (0.2, 0.836)
    (0.3, 0.845)
    (0.5, 0.856)
    (1.0, 0.867)
    };
    \addlegendentry{Mixture} 
\end{axis}
\end{tikzpicture}
\caption{Accuracy of \textit{Incremental} on TabMWP with value networks trained with \textit{Single} or \textit{Mixture} domains data, where we use full GSM8K and different ratio of TabMWP from 1\% to 100\%.}
\label{fig:domain_mix}
\end{figure}
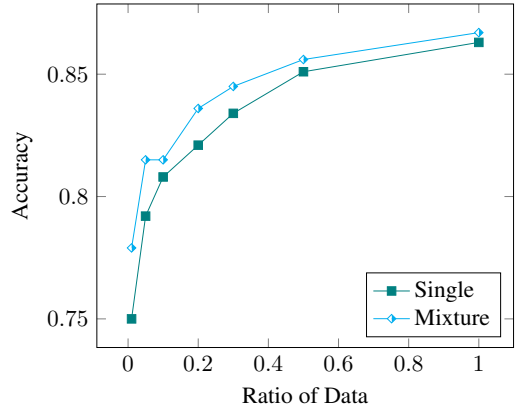

\paragraph{\textit{Mixture-of-Domain Boosts Performance}}
An important future direction is to construct a general value network that can address questions from different domains.
To validate the potential of this direction, we conducted experiments using value networks trained with different ratios of TabMWP data and full GSM8K data.
Despite the significant difference in question style and answer type, the results in Fig. \ref{fig:domain_mix} demonstrate that using a mixture of different domains helps improve search performance, especially when training instances from the target domain are scarce (0.75 vs. 0.78 on Accuracy when using 1\% TabMWP data).
This highlights the effectiveness of building robust and stronger value networks by collecting various training instances.
Further exploration in this direction will be pursued in future work.

\subsection{Limitations of Guided Tree Search}
\label{sec:error}

Analyses above have shown the effectiveness of our method.
However, though much more efficient, guided tree searches often cannot outperform Soft Voting on Accuracy.
This section will provide detailed analyses to answer this question.

\begin{figure}[t]
    \includegraphics[width=1.\linewidth]{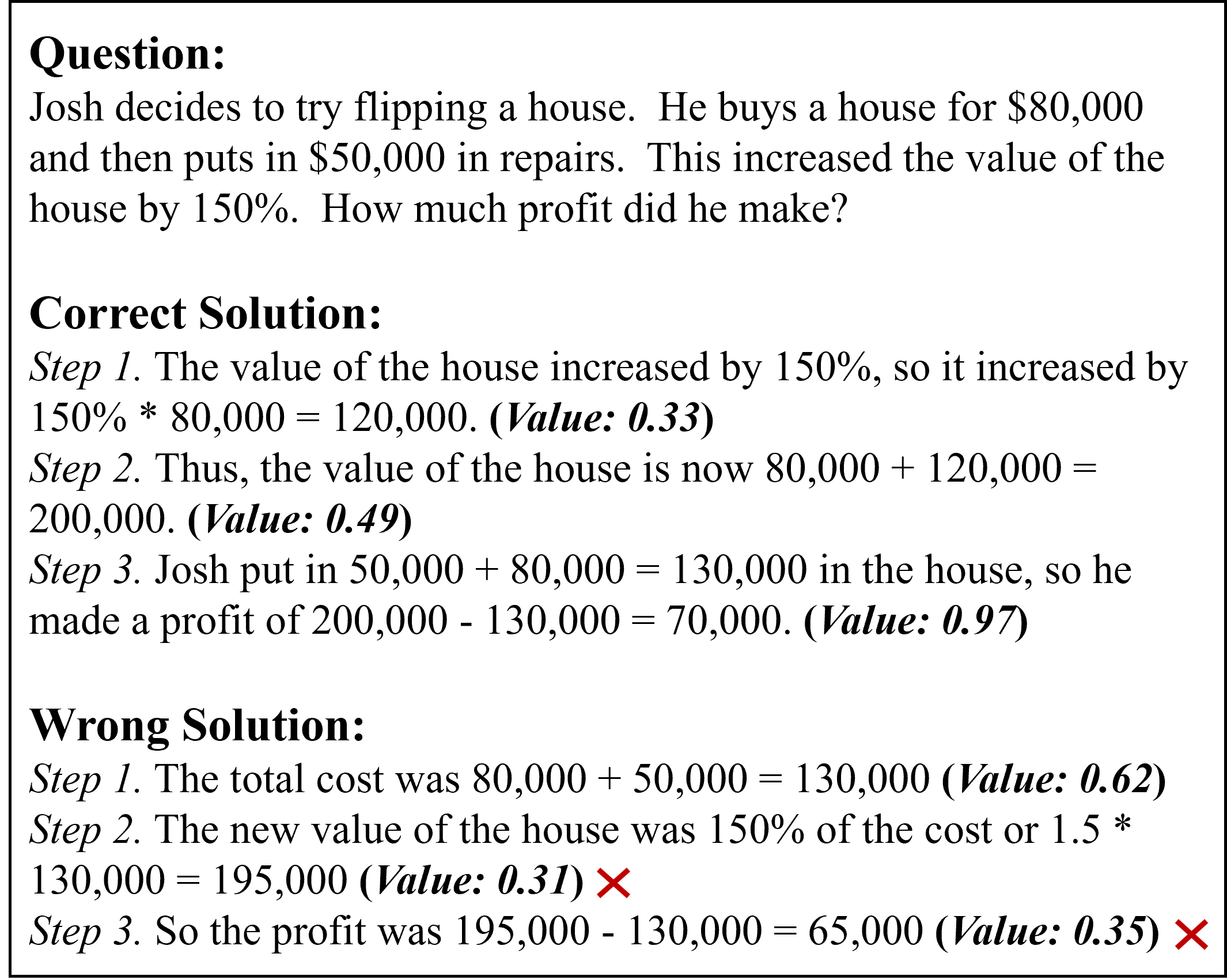}
    \caption{A lemon picked example. We provide correct and wrong solutions with value scores for each step and error steps are marked.}
    \label{fig:case_study}
\end{figure}

\paragraph{\textit{Larger Variance of Values at the First Few Steps}}
We first collect the questions that our approach fails to solve, yet are successfully addressed by \textit{Soft Voting@20}.
Fig. \ref{fig:case_study} displays a lemon pick example.
We observe that our value network can select the correct answer when the complete rationale is provided.
However, the first two steps in the correct path are scored much lower than the first step of the wrong solution.
This results in a reduced priority in exploring the node that is actually of higher quality.
We also compute the average best ranking of values for correct solutions across the 20 sampled paths for these questions.
Results indicate that the rank for the first step is 3.0, whereas the final step achieves a rank of 1.4.
This finding highlights the larger variance of values at the first few steps, which subsequently results in inadequate exploration of high-quality nodes and finally fails to yield the correct answer.

\begin{table}[]
\small
    \centering
    \begin{tabular}{lcc}
    \toprule
    & Accuracy $\uparrow$ & Tokens (\emph{k}) $\downarrow$ \\
    \midrule
    Best@20 & .833 & 2.63 \\
    Soft Voting@20 & .843 & 2.63 \\
    \hdashline
    Ours (Batch) & .823 & \textbf{0.55} \\
    ~~~~$+$Soft Voting ($\alpha = 0.7$) & .841 & 0.73 \\
    ~~~~$+$Soft Voting ($\alpha = 0.8$) & .843 & 0.79 \\
    ~~~~$+$Soft Voting ($\alpha = 0.9$) & \textbf{.847} & 0.99 \\
    \bottomrule
    \end{tabular}
    \caption{Results of our methods enhanced by voting, where \textit{Best@20} is a variant of \textit{Soft Voting@20}, which only selects the best path with the highest value as the final output. For \textit{Ours + Soft Voting}, we discard results with values lower than $\alpha$, and utilize \textit{Soft Voting@20} to solve them.}
    \label{tab:plus_vote}
\end{table}

\paragraph{\textit{Voting Helps when Values are Inaccurate}}
Due to the imperfect value network, some incorrect paths may be erroneously scored higher than the correct ones.
Guided tree search methods, which search for only one path as the final answer, inevitably fail in these instances.
However, Soft Voting can mitigate this issue by leveraging the benefits of both majority voting and the value network.
Consequently, even if the highest value is attained by an incorrect path, Soft Voting still has the potential to reach the correct answer with a higher frequency.
As demonstrated in Table \ref{tab:plus_vote}, the use of voting enables \textit{Soft Voting@20} to outperform \textit{Best@20}, highlighting the efficacy of voting in enhancing accuracy.

Inspired by these findings, we further investigate the improvement of our method using voting.
Specifically, we discard the answers predicted by our method when their value scores fall below a threshold $\alpha$.
Generally, these predictions exhibit a higher error rate due to the correlation between value scores and correctness.
Subsequently, we employ Soft Voting to address these unresolved questions.
The results in Table \ref{tab:plus_vote} indicate that accuracy can be significantly improved by increasing $\alpha$.
However, the associated costs also rise substantially, albeit remaining lower than those of \textit{Soft Voting@20}.

\section{Conclusion}
In this work, we study guided tree search to address math problems, aiming to decrease the computation costs while maintaining the performance.
Inspired by the theory of value function, 
we propose dynamic node selection and expansion strategies, which dynamically determine the priority of nodes to explore and manage the computational budget during expansion.
Both procedures are guided by an easy-to-implement value network trained without step-wise supervision.
Experiments show that our methods achieve competitive performance with typical baselines but significantly save computation costs.
Ablation studies validate the effectiveness of each component, providing more feasible options for various practical scenarios.
Besides, we also identify the shortcomings of this research line, and provide a potential strategy for addressing these issues.

% \section*{Limitations}
% xxx

\bibliography{custom}

\begin{thebibliography}{41}
\providecommand{\natexlab}[1]{#1}

\bibitem[{Achiam et~al.(2023)Achiam, Adler, Agarwal, Ahmad, Akkaya, Aleman, Almeida, Altenschmidt, Altman, Anadkat et~al.}]{achiam2023gpt}
Josh Achiam, Steven Adler, Sandhini Agarwal, Lama Ahmad, Ilge Akkaya, Florencia~Leoni Aleman, Diogo Almeida, Janko Altenschmidt, Sam Altman, Shyamal Anadkat, et~al. 2023.
\newblock Gpt-4 technical report.
\newblock \emph{arXiv preprint arXiv:2303.08774}.

\bibitem[{Amini et~al.(2019)Amini, Gabriel, Lin, Koncel-Kedziorski, Choi, and Hajishirzi}]{amini2019mathqa}
Aida Amini, Saadia Gabriel, Shanchuan Lin, Rik Koncel-Kedziorski, Yejin Choi, and Hannaneh Hajishirzi. 2019.
\newblock Mathqa: Towards interpretable math word problem solving with operation-based formalisms.
\newblock In \emph{Proceedings of the 2019 Conference of the North American Chapter of the Association for Computational Linguistics: Human Language Technologies, Volume 1 (Long and Short Papers)}, pages 2357--2367.

\bibitem[{Chen et~al.(2024)Chen, Deng, Yuan, Ji, and Gu}]{chen2024self}
Zixiang Chen, Yihe Deng, Huizhuo Yuan, Kaixuan Ji, and Quanquan Gu. 2024.
\newblock Self-play fine-tuning converts weak language models to strong language models.
\newblock \emph{arXiv preprint arXiv:2401.01335}.

\bibitem[{Cobbe et~al.(2021)Cobbe, Kosaraju, Bavarian, Chen, Jun, Kaiser, Plappert, Tworek, Hilton, Nakano et~al.}]{cobbe2021training}
Karl Cobbe, Vineet Kosaraju, Mohammad Bavarian, Mark Chen, Heewoo Jun, Lukasz Kaiser, Matthias Plappert, Jerry Tworek, Jacob Hilton, Reiichiro Nakano, et~al. 2021.
\newblock Training verifiers to solve math word problems.
\newblock \emph{arXiv preprint arXiv:2110.14168}.

\bibitem[{Daniel(2017)}]{daniel2017thinking}
Kahneman Daniel. 2017.
\newblock \emph{Thinking, fast and slow}.

\bibitem[{Feng et~al.(2023)Feng, Wan, Wen, Wen, Zhang, and Wang}]{feng2023alphazero}
Xidong Feng, Ziyu Wan, Muning Wen, Ying Wen, Weinan Zhang, and Jun Wang. 2023.
\newblock Alphazero-like tree-search can guide large language model decoding and training.
\newblock \emph{arXiv preprint arXiv:2309.17179}.

\bibitem[{Hendrycks et~al.(2021)Hendrycks, Burns, Kadavath, Arora, Basart, Tang, Song, and Steinhardt}]{hendrycks2021measuring}
Dan Hendrycks, Collin Burns, Saurav Kadavath, Akul Arora, Steven Basart, Eric Tang, Dawn Song, and Jacob Steinhardt. 2021.
\newblock Measuring mathematical problem solving with the math dataset.
\newblock In \emph{Thirty-fifth Conference on Neural Information Processing Systems Datasets and Benchmarks Track (Round 2)}.

\bibitem[{Hopkins et~al.(2017)Hopkins, Petrescu-Prahova, Levin, Le~Bras, Herrasti, and Joshi}]{hopkins2017beyond}
Mark Hopkins, Cristian Petrescu-Prahova, Roie Levin, Ronan Le~Bras, Alvaro Herrasti, and Vidur Joshi. 2017.
\newblock Beyond sentential semantic parsing: Tackling the math sat with a cascade of tree transducers.
\newblock In \emph{Proceedings of the 2017 Conference on Empirical Methods in Natural Language Processing}, pages 795--804.

\bibitem[{Jiang et~al.(2024)Jiang, Sablayrolles, Roux, Mensch, Savary, Bamford, Chaplot, Casas, Hanna, Bressand et~al.}]{jiang2024mixtral}
Albert~Q Jiang, Alexandre Sablayrolles, Antoine Roux, Arthur Mensch, Blanche Savary, Chris Bamford, Devendra~Singh Chaplot, Diego de~las Casas, Emma~Bou Hanna, Florian Bressand, et~al. 2024.
\newblock Mixtral of experts.
\newblock \emph{arXiv preprint arXiv:2401.04088}.

\bibitem[{Kang et~al.(2024)Kang, Li, Chen, Kazemi, and Chen}]{kang2024mindstar}
Jikun Kang, Xin~Zhe Li, Xi~Chen, Amirreza Kazemi, and Boxing Chen. 2024.
\newblock Mindstar: Enhancing math reasoning in pre-trained llms at inference time.
\newblock \emph{arXiv preprint arXiv:2405.16265}.

\bibitem[{Khalifa et~al.(2023)Khalifa, Logeswaran, Lee, Lee, and Wang}]{khalifa2023grace}
Muhammad Khalifa, Lajanugen Logeswaran, Moontae Lee, Honglak Lee, and Lu~Wang. 2023.
\newblock Grace: Discriminator-guided chain-of-thought reasoning.
\newblock In \emph{Findings of the Association for Computational Linguistics: EMNLP 2023}, pages 15299--15328.

\bibitem[{Kojima et~al.(2022)Kojima, Gu, Reid, Matsuo, and Iwasawa}]{kojima2022large}
Takeshi Kojima, Shixiang~Shane Gu, Machel Reid, Yutaka Matsuo, and Yusuke Iwasawa. 2022.
\newblock Large language models are zero-shot reasoners.
\newblock \emph{Advances in neural information processing systems}, 35:22199--22213.

\bibitem[{Li et~al.(2019)Li, Wang, Zhang, Wang, Dai, and Zhang}]{li2019modeling}
Jierui Li, Lei Wang, Jipeng Zhang, Yan Wang, Bing~Tian Dai, and Dongxiang Zhang. 2019.
\newblock Modeling intra-relation in math word problems with different functional multi-head attentions.
\newblock In \emph{Proceedings of the 57th annual meeting of the association for computational linguistics}, pages 6162--6167.

\bibitem[{Lightman et~al.(2023)Lightman, Kosaraju, Burda, Edwards, Baker, Lee, Leike, Schulman, Sutskever, and Cobbe}]{lightman2023let}
Hunter Lightman, Vineet Kosaraju, Yura Burda, Harri Edwards, Bowen Baker, Teddy Lee, Jan Leike, John Schulman, Ilya Sutskever, and Karl Cobbe. 2023.
\newblock Let's verify step by step.
\newblock \emph{arXiv preprint arXiv:2305.20050}.

\bibitem[{Liu and Yao(2024)}]{liu2024augmenting}
Haoxiong Liu and Andrew Chi-Chih Yao. 2024.
\newblock Augmenting math word problems via iterative question composing.
\newblock \emph{arXiv preprint arXiv:2401.09003}.

\bibitem[{Loshchilov and Hutter(2017)}]{loshchilov2017decoupled}
Ilya Loshchilov and Frank Hutter. 2017.
\newblock Decoupled weight decay regularization.
\newblock \emph{arXiv preprint arXiv:1711.05101}.

\bibitem[{Lu et~al.(2022)Lu, Qiu, Chang, Wu, Zhu, Rajpurohit, Clark, and Kalyan}]{lu2022dynamic}
Pan Lu, Liang Qiu, Kai-Wei Chang, Ying~Nian Wu, Song-Chun Zhu, Tanmay Rajpurohit, Peter Clark, and Ashwin Kalyan. 2022.
\newblock Dynamic prompt learning via policy gradient for semi-structured mathematical reasoning.
\newblock In \emph{The Eleventh International Conference on Learning Representations}.

\bibitem[{Luo et~al.(2023)Luo, Sun, Xu, Zhao, Lou, Tao, Geng, Lin, Chen, and Zhang}]{luo2023wizardmath}
Haipeng Luo, Qingfeng Sun, Can Xu, Pu~Zhao, Jianguang Lou, Chongyang Tao, Xiubo Geng, Qingwei Lin, Shifeng Chen, and Dongmei Zhang. 2023.
\newblock Wizardmath: Empowering mathematical reasoning for large language models via reinforced evol-instruct.
\newblock \emph{arXiv preprint arXiv:2308.09583}.

\bibitem[{Ma et~al.(2023)Ma, Zhou, Liu, Yuan, Liu, You, and Yang}]{ma2023let}
Qianli Ma, Haotian Zhou, Tingkai Liu, Jianbo Yuan, Pengfei Liu, Yang You, and Hongxia Yang. 2023.
\newblock Let's reward step by step: Step-level reward model as the navigators for reasoning.
\newblock \emph{arXiv preprint arXiv:2310.10080}.

\bibitem[{Madaan et~al.(2024)Madaan, Tandon, Gupta, Hallinan, Gao, Wiegreffe, Alon, Dziri, Prabhumoye, Yang et~al.}]{madaan2024self}
Aman Madaan, Niket Tandon, Prakhar Gupta, Skyler Hallinan, Luyu Gao, Sarah Wiegreffe, Uri Alon, Nouha Dziri, Shrimai Prabhumoye, Yiming Yang, et~al. 2024.
\newblock Self-refine: Iterative refinement with self-feedback.
\newblock \emph{Advances in Neural Information Processing Systems}, 36.

\bibitem[{Matsuzaki et~al.(2017)Matsuzaki, Ito, Iwane, Anai, and Arai}]{matsuzaki2017semantic}
Takuya Matsuzaki, Takumi Ito, Hidenao Iwane, Hirokazu Anai, and Noriko~H Arai. 2017.
\newblock Semantic parsing of pre-university math problems.
\newblock In \emph{Proceedings of the 55th Annual Meeting of the Association for Computational Linguistics (Volume 1: Long Papers)}, pages 2131--2141.

\bibitem[{Qin et~al.(2021)Qin, Liang, Hong, Tang, and Lin}]{qin2021neural}
Jinghui Qin, Xiaodan Liang, Yining Hong, Jianheng Tang, and Liang Lin. 2021.
\newblock Neural-symbolic solver for math word problems with auxiliary tasks.
\newblock In \emph{Proceedings of the 59th Annual Meeting of the Association for Computational Linguistics and the 11th International Joint Conference on Natural Language Processing (Volume 1: Long Papers)}, pages 5870--5881.

\bibitem[{Sutton and Barto(2018)}]{sutton2018reinforcement}
Richard~S Sutton and Andrew~G Barto. 2018.
\newblock \emph{Reinforcement learning: An introduction}.
\newblock MIT press.

\bibitem[{Suzgun and Kalai(2024)}]{suzgun2024meta}
Mirac Suzgun and Adam~Tauman Kalai. 2024.
\newblock Meta-prompting: Enhancing language models with task-agnostic scaffolding.
\newblock \emph{arXiv preprint arXiv:2401.12954}.

\bibitem[{Tian et~al.(2024)Tian, Peng, Song, Jin, Yu, Mi, and Yu}]{tian2024toward}
Ye~Tian, Baolin Peng, Linfeng Song, Lifeng Jin, Dian Yu, Haitao Mi, and Dong Yu. 2024.
\newblock Toward self-improvement of llms via imagination, searching, and criticizing.
\newblock \emph{arXiv preprint arXiv:2404.12253}.

\bibitem[{Toshniwal et~al.(2024)Toshniwal, Moshkov, Narenthiran, Gitman, Jia, and Gitman}]{toshniwal2024openmathinstruct}
Shubham Toshniwal, Ivan Moshkov, Sean Narenthiran, Daria Gitman, Fei Jia, and Igor Gitman. 2024.
\newblock Openmathinstruct-1: A 1.8 million math instruction tuning dataset.
\newblock \emph{arXiv preprint arXiv:2402.10176}.

\bibitem[{Touvron et~al.(2023)Touvron, Martin, Stone, Albert, Almahairi, Babaei, Bashlykov, Batra, Bhargava, Bhosale et~al.}]{touvron2023llama}
Hugo Touvron, Louis Martin, Kevin Stone, Peter Albert, Amjad Almahairi, Yasmine Babaei, Nikolay Bashlykov, Soumya Batra, Prajjwal Bhargava, Shruti Bhosale, et~al. 2023.
\newblock Llama 2: Open foundation and fine-tuned chat models.
\newblock \emph{arXiv preprint arXiv:2307.09288}.

\bibitem[{Wang et~al.(2024{\natexlab{a}})Wang, Song, Tian, Peng, Jin, Mi, Su, and Yu}]{wang2024self}
Ante Wang, Linfeng Song, Ye~Tian, Baolin Peng, Lifeng Jin, Haitao Mi, Jinsong Su, and Dong Yu. 2024{\natexlab{a}}.
\newblock Self-consistency boosts calibration for math reasoning.
\newblock \emph{arXiv preprint arXiv:2403.09849}.

\bibitem[{Wang et~al.(2024{\natexlab{b}})Wang, Deng, Lv, Yan, and Bo}]{wang2024q}
Chaojie Wang, Yanchen Deng, Zhiyi Lv, Shuicheng Yan, and An~Bo. 2024{\natexlab{b}}.
\newblock Q*: Improving multi-step reasoning for llms with deliberative planning.
\newblock \emph{arXiv preprint arXiv:2406.14283}.

\bibitem[{Wang et~al.(2023)Wang, Li, Shao, Xu, Dai, Li, Chen, Wu, and Sui}]{wang2023math}
Peiyi Wang, Lei Li, Zhihong Shao, RX~Xu, Damai Dai, Yifei Li, Deli Chen, Y~Wu, and Zhifang Sui. 2023.
\newblock Math-shepherd: A label-free step-by-step verifier for llms in mathematical reasoning.
\newblock \emph{arXiv preprint arXiv:2312.08935}.

\bibitem[{Wang et~al.(2022)Wang, Wei, Schuurmans, Le, Chi, Narang, Chowdhery, and Zhou}]{wang2022self}
Xuezhi Wang, Jason Wei, Dale Schuurmans, Quoc~V Le, Ed~H Chi, Sharan Narang, Aakanksha Chowdhery, and Denny Zhou. 2022.
\newblock Self-consistency improves chain of thought reasoning in language models.
\newblock In \emph{The Eleventh International Conference on Learning Representations}.

\bibitem[{Wei et~al.(2022)Wei, Wang, Schuurmans, Bosma, Xia, Chi, Le, Zhou et~al.}]{wei2022chain}
Jason Wei, Xuezhi Wang, Dale Schuurmans, Maarten Bosma, Fei Xia, Ed~Chi, Quoc~V Le, Denny Zhou, et~al. 2022.
\newblock Chain-of-thought prompting elicits reasoning in large language models.
\newblock \emph{Advances in neural information processing systems}, 35:24824--24837.

\bibitem[{Wu et~al.(2021)Wu, Zhang, Wei, and Huang}]{wu2021math}
Qinzhuo Wu, Qi~Zhang, Zhongyu Wei, and Xuan-Jing Huang. 2021.
\newblock Math word problem solving with explicit numerical values.
\newblock In \emph{Proceedings of the 59th Annual Meeting of the Association for Computational Linguistics and the 11th International Joint Conference on Natural Language Processing (Volume 1: Long Papers)}, pages 5859--5869.

\bibitem[{Xie et~al.(2024)Xie, Kawaguchi, Zhao, Zhao, Kan, He, and Xie}]{xie2024self}
Yuxi Xie, Kenji Kawaguchi, Yiran Zhao, James~Xu Zhao, Min-Yen Kan, Junxian He, and Michael Xie. 2024.
\newblock Self-evaluation guided beam search for reasoning.
\newblock \emph{Advances in Neural Information Processing Systems}, 36.

\bibitem[{Yang et~al.(2024)Yang, Wang, Lu, Liu, Le, Zhou, and Chen}]{yang2024largelanguagemodelsoptimizers}
Chengrun Yang, Xuezhi Wang, Yifeng Lu, Hanxiao Liu, Quoc~V. Le, Denny Zhou, and Xinyun Chen. 2024.
\newblock \href {https://arxiv.org/abs/2309.03409} {Large language models as optimizers}.
\newblock \emph{Preprint}, arXiv:2309.03409.

\bibitem[{Yao et~al.(2024)Yao, Yu, Zhao, Shafran, Griffiths, Cao, and Narasimhan}]{yao2024tree}
Shunyu Yao, Dian Yu, Jeffrey Zhao, Izhak Shafran, Tom Griffiths, Yuan Cao, and Karthik Narasimhan. 2024.
\newblock Tree of thoughts: Deliberate problem solving with large language models.
\newblock \emph{Advances in Neural Information Processing Systems}, 36.

\bibitem[{Yu et~al.(2023)Yu, Jiang, Shi, Jincheng, Liu, Zhang, Kwok, Li, Weller, and Liu}]{yu2023metamath}
Longhui Yu, Weisen Jiang, Han Shi, YU~Jincheng, Zhengying Liu, Yu~Zhang, James Kwok, Zhenguo Li, Adrian Weller, and Weiyang Liu. 2023.
\newblock Metamath: Bootstrap your own mathematical questions for large language models.
\newblock In \emph{The Twelfth International Conference on Learning Representations}.

\bibitem[{Zeng et~al.(2023)Zeng, Watson, Cho, Rahimi, Reynolds, Balch, and Veloso}]{zeng2023flowmind}
Zhen Zeng, William Watson, Nicole Cho, Saba Rahimi, Shayleen Reynolds, Tucker Balch, and Manuela Veloso. 2023.
\newblock Flowmind: automatic workflow generation with llms.
\newblock In \emph{Proceedings of the Fourth ACM International Conference on AI in Finance}, pages 73--81.

\bibitem[{Zhang et~al.(2024)Zhang, Li, Huang, Zhou, Li, and Ouyang}]{zhang2024accessing}
Di~Zhang, Jiatong Li, Xiaoshui Huang, Dongzhan Zhou, Yuqiang Li, and Wanli Ouyang. 2024.
\newblock Accessing gpt-4 level mathematical olympiad solutions via monte carlo tree self-refine with llama-3 8b.
\newblock \emph{arXiv preprint arXiv:2406.07394}.

\bibitem[{Zhou et~al.(2024)Zhou, Pujara, Ren, Chen, Cheng, Le, Chi, Zhou, Mishra, and Zheng}]{zhou2024self}
Pei Zhou, Jay Pujara, Xiang Ren, Xinyun Chen, Heng-Tze Cheng, Quoc~V Le, Ed~H Chi, Denny Zhou, Swaroop Mishra, and Huaixiu~Steven Zheng. 2024.
\newblock Self-discover: Large language models self-compose reasoning structures.
\newblock \emph{arXiv preprint arXiv:2402.03620}.

\bibitem[{Zhu et~al.(2024)Zhu, Zhang, Xie, and Su}]{zhu2024deductive}
Tinghui Zhu, Kai Zhang, Jian Xie, and Yu~Su. 2024.
\newblock Deductive beam search: Decoding deducible rationale for chain-of-thought reasoning.
\newblock \emph{arXiv preprint arXiv:2401.17686}.

\end{thebibliography}

\appendix

\begin{figure}[t]
    \includegraphics[width=1.\linewidth]{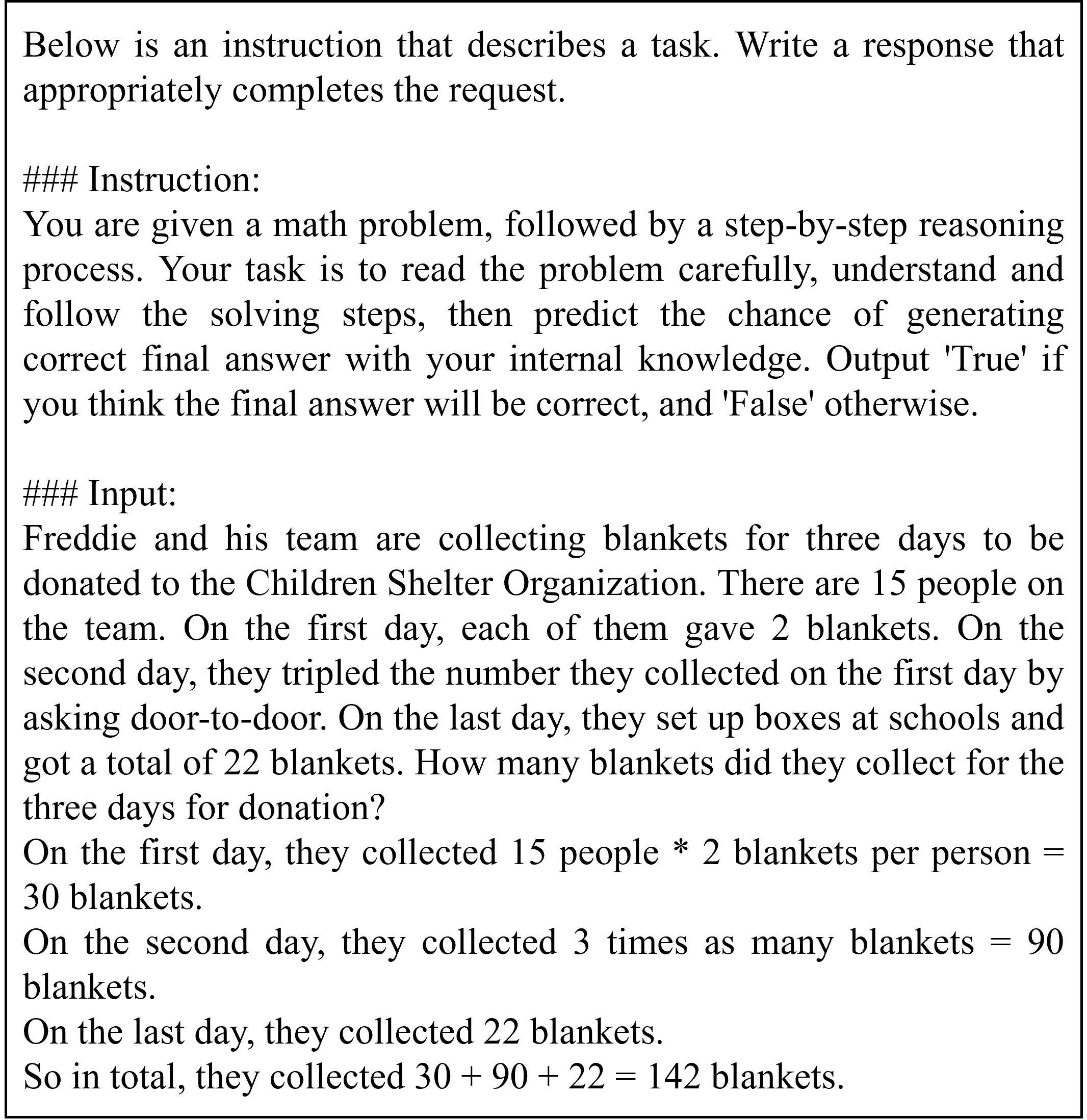}
    \caption{An example fed to value network for scoring.}
    \label{fig:instruction}
\end{figure}

% \section{Example Appendix}
% \label{sec:appendix}

% This is an appendix.

\end{document}